\documentclass[10pt,twocolumn,letterpaper]{article}

\usepackage{iccv}
\usepackage{times}
\usepackage{epsfig}
\usepackage{graphicx}
\usepackage{amsmath}
\usepackage{amssymb}
\usepackage{color}
\usepackage[table]{xcolor} 
\usepackage{ifthen}
\usepackage{multirow}

\usepackage[breaklinks=true,bookmarks=false,colorlinks = true]{hyperref}

 \iccvfinalcopy 

\def\iccvPaperID{****} 

\makeatletter
\newcommand*\titleheader[1]{\gdef\@titleheader{#1}}
\AtBeginDocument{
  \let\st@red@title\@title
  \def\@title{%
    \bgroup\normalfont\large\centering\@titleheader\par\egroup
    \vskip1.5em\st@red@title}
}
\makeatother

\title{Texture and Structure Incorporated ScatterNet Hybrid Deep Learning Network (TS-SHDL) For Brain Matter Segmentation}
\titleheader{To Appear in the IEEE International Conference on Computer Vision Workshops (ICCVW) 2017}

\begin{document}


\author{Amarjot Singh\\
Signal Processing Group \\ 
Department of Engineering \\
University of Cambridge, U.K.\\
{\tt\small as2436@cam.ac.uk}
\and
Devamanyu Hazarika\\
School of Computing \\
National University of Singapore \\
Singapore \\
{\tt\small devamanyu@comp.nus.edu.sg}
\and
Aniruddha Bhattacharya\\
Department of Computer Science \\
National Institute of Technology \\
Warangal, India \\
{\tt\small baniruddha2@student.nitw.ac.in}}

\maketitle

\begin{abstract}
Automation of brain matter segmentation from MR images is a challenging task due to the irregular boundaries between the grey and white matter regions. In addition, the presence of intensity inhomogeneity in the MR images further complicates the problem. In this paper, we propose a texture and vesselness incorporated version of the ScatterNet Hybrid Deep Learning Network (TS-SHDL) that extracts hierarchical invariant mid-level features, used by fisher vector encoding and a conditional random field (CRF) to perform the desired segmentation. The performance of the proposed network is evaluated by extensive experimentation and comparison with the state-of-the-art methods on several 2D MRI scans taken from the synthetic McGill Brain Web as well as on the MRBrainS dataset of real 3D MRI scans. The advantages of the TS-SHDL network over supervised deep learning networks is also presented in addition to its superior performance over the state-of-the-art.
\end{abstract}

\section{Introduction}
Brain matter segmentation is an important task that is essential for the study of various ailments like Alzheimer and Parkinson disease~\cite{lowe2004distinctive} etc. It is typically done by skilled professionals who manually label white and grey matter in the Magnetic Resonance (MR) images of the brain. The images also often contain noise incurred through acquisition defects and errors~\cite{vovk2007review} that further adds to the complexity of the problem. Manual labeling leads to errors and inconsistencies thus requiring the need for automatic labeling. This task has been performed in three different ways over the years.

The first class of methods utilized hand-engineered features extracted using filters designed to capture edge and texture representations~\cite{kapur1996segmentation,angoth2013novel,dwith2013wavelet,amar1}. For example, SIFT~\cite{lowe} features were used in the application of reconstructing tractograms of whole brain~\cite{smith2013sift}. These hand-engineered features are easy to design but are constrained by marginally good performance. 

\begin{figure*}[t!]
	\centering
	\includegraphics[width=1\textwidth]{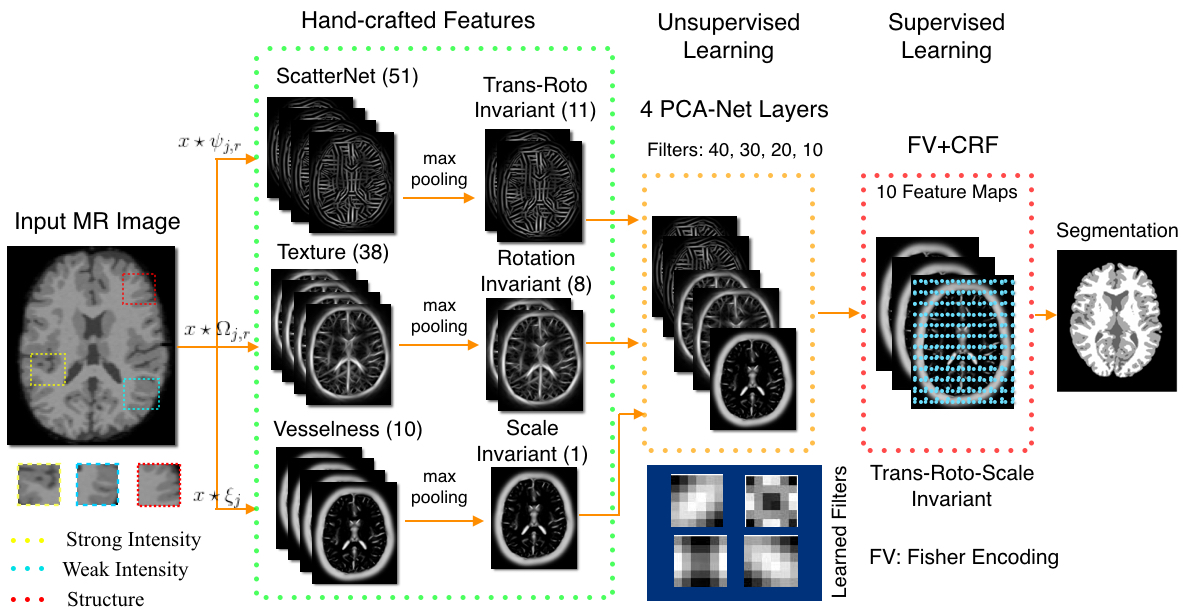}
	\caption{The Illustration shows the proposed TS-SHDL network that captures Intensity (region boundaries and texture) and structural (ridge like structures) properties of the brain matter with handcrafted (ScatterNet, Vesselness, and Texture) features which are concatenated and used by four stacked layer of the PCANet to learn 40, 30, 20 and 10 filters at the first, second, third and fourth layers respectively, of size 5 $\times$ 5. The filters capture a hierarchy of translation, rotation, and scale-invariant mid-level features. The invariant mid-level features are encoded into a dense and compact representation Fisher vector encoding of 500 dimensions using 5 Gaussian mixtures. The encoded features are used by the CRF to produce the desired segmentation.}
	\label{fig:properties}
\end{figure*}

The second class of methods used deep networks for brain matter segmentation as demonstrated in multiple works \cite{bezdek1993review,havaei2017brain}. Deep networks have achieved the state-of-the-art performance by learning discriminative class-specific features. However, deep networks suffer from overwhelming complexity of trainable parameters and design of optimal configurations. Thus, training with limited sized datasets, common in medical imaging, still remains a difficult task. 

The third class of methods used hand-crafted features to first extract low-level edge features which are further used by supervised or unsupervised learning methods to learn higher levels of representations, thus resulting in a hybrid network. For example, Zhang et al.~\cite{zhang2015deep} extracted anisotropic features which were used by a CNN to learn higher-level representations. Given the reduced parameter set to be learned while training, these networks generally perform better than normal deep neural networks on small datasets with less training instance, a trait always seen in brain segmentation datasets. 

Singh et.~al.~\cite{shdl} recently proposed a ScatterNet Hybrid Deep Learning (SHDL) network that first extracts ScatterNet handcrafted low-level descriptors which are used by two layers of PCA-Net based unsupervised learning module to learn hierarchical object specific mid-level features. The mid-level features are finally used by a supervised module composed of orthogonal least squares and support vector machine (SVM) to perform classification. Each layer of the SHDL network is designed and optimized automatically using cross-validation that produces a computationally efficient architecture.

This paper proposes an improved texture and vesselness incorporated ScatterNet Hybrid Deep Learning (TS-SHDL) network for brain matter segmentation in the MR images. The improved network incorporates the vesselness and texture hand-crafted features in addition to the two-layer ScatterNet descriptors to capture the intensity and structural properties of brain matter as shown in Fig. 1. The extracted hand-crafted features are used by four stacked layers of unsupervised learning based PCA-Net to learn translation, rotation and scale invariant mid-level features. The mid-level features are used by the supervised Fisher encoding (FK) to gradually reduce the dimensions of the feature vector. The reduced feature vector is finally used by a conditional random field (CRF) to perform the desired brain matter segmentation. The contributions and reasoning for the choosing the above-mentioned handcrafted features, unsupervised, and supervised learning, module are discussed below. 

 The main contributions of the paper are stated below:
\begin{itemize}

\item \textit{Hand-crafted Module}: Texture and Veselness feature are used jointly with ScatterNet hand-crafted descriptors as they capture the intensity and structural properties of the brain matter. The intensity properties are represented in the brain matter by region boundaries and texture while the ridge-like structures correspond to the structural properties of the matter as shown in Fig. 1. Parametric DTCWT ScatterNet~\cite{singh2017dual} and Texture \cite{pitiot2004expert} filters are used to represent the intensity properties of the brain while the structural properties were extracted using a vesselness \cite{lindeberg1998edge} filter. The intensity and structural properties can appear at any position,
orientation and scale in the MR images. Hence, invariance to translation, rotation, and scale is introduced in each sub-feature. 

\item \textit{Unsupervised Learning Module}: This module uses four stacked PCA-Net~\cite{pcanet} layers on the concatenated hand-crafted features to learn translation, rotation, and scale invariant robust hierarchical mid-level features. The network is fast and easy to train as opposed to other unsupervised learning modules such autoencoders or RBMs as the minimization of the loss function (Eq. 6) can be obtained in its simplistic form as
the eigen decomposition. This makes PCA-Net a good
method for learning mid-level features.

\item \textit{Supervised Learning Module}: The features obtained from the PCA-Net module are reduced to a more compact and dense representation using Fisher Encoding (FV)~\cite{FV}. The reduced dimensions results in faster learning of the Conditional Random Field (CRF), used to perform the desired segmentation. 

\item \textit{Advantages over Supervised Deep Networks}: The proposed network makes use of unsupervised learning to learn a hierarchy of mid-level features that are utilized to perform the desired brain matter segmentation. Similar hierarchical features can be also learned using supervised deep networks. However, supervised deep networks requires large amounts of annotated training data to learn these features which may not available for many applications~~\cite{satellite,su,medical}, especially for medical image applications. The proposed TS-SHDL network can be a potential alternative to deep networks for applications with small training datasets. 

  \end{itemize}

The proposed framework is used to perform brain matter segmentation on 2D synthetic and 3D real brain MRI datasets. The average segmentation accuracy for each class for both datasets is presented. In addition, an extensive comparison of the proposed pipeline with other deep brain matter segmentation methods is demonstrated.

The paper is divided into the following sections. Section 2 briefly presents the proposed TS-SHDL network. Section 3 presents the experimental results while Section 4 draws conclusions.

\section{TS-SHDL Network}
This section details the texture and structure feature incorporated ScatterNet Hybrid Deep Learning Network (TS-SHDL) used for brain matter segmentation. The first sub-section explains the mathematical formulation of the ScatterNet, texture and structure, sub-features, combined to form the hand-crafted features module. The second sub-section details the mathematical formulation of the four layers unsupervised learning based PCA-Net framework that utilizes the hand-crafted features to extract translation, rotation, and scale invariant features. The final sub-section details the supervised Fisher vector encoding that produces a compact and dense feature vector finally used by the Conditional Random Field to produce the desired brain matter segmentation. The TS-SHDL network is presented in Fig. 1.

\subsection{Hand-crafted Features}
This section details the hand-crafted features extracted from the MR images that capture the intensity and structure features as shown in Fig. 1.

\subsubsection{DTCWT ScatterNet}
The parametric log based DTCWT ScatterNet~\cite{singh2017dual} is an improved version (both on classification error and computational efficiency) of the multi-layer Scattering Network~\cite{Jbruna2013,Oyallon2015,ima,eccv} that extracts relatively symmetric translation invariant representations from a multi-resolution image using the \textit{dual-tree complex wavelet transform} (DTCWT)~\cite{Kingsbury1998} and parametric log transformation layer. Below we present the formulation of the parametric DTCWT ScatterNet for a single input image which may then be applied to each of the multi-resolution images.

The first layer is involved in filtering the input signal $x$ using dual-tree complex wavelets $ \psi_{j,r }$ at different scales ($j$). Six pre-defined orientations are fixed for this operation -  $15^\circ, 45^\circ, 75^\circ, 105^\circ, 135^\circ$ and $165^\circ$. Point-wise $L_{2}$ non-linearity is applied to the filtered signal, to build more translation invariant representation as described below:

\begin{figure*}[t!]
	\centering
	\includegraphics[width = 1\textwidth]{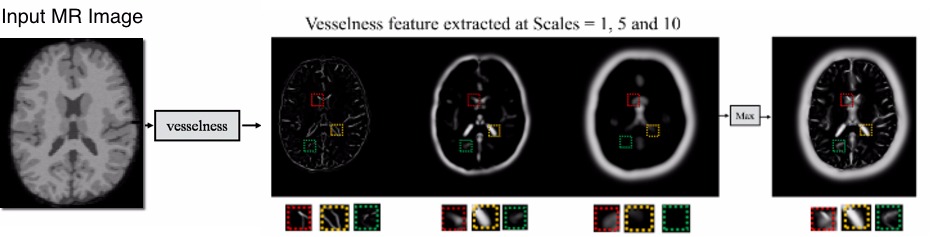}
	\caption{Demonstrates the pipeline of vesselness features at 3 scales. For each feature map extracted at each scale, a $20 \times 20$ patch is selected. Here only 3 scales are shown as opposed to 10 used in our experiments owing to space restrictions. Finally, Max operation is performed across scales to achieve scale invariance.}
	\label{fig:vesselness}
\end{figure*}

\begin{equation}
U[\lambda_{m = 1}] = |x\star \psi_{\lambda_{1} }| = \sqrt{|x\star \psi_{\lambda_{1} }^{a}|^2 + |x\star \psi_{\lambda_{1} }^{b}|^2} 
\end{equation}
After this a parametric log transformation layer is applied to all the oriented representations ($U[j]$). This extracts representations at a particular scale $j$ with a parameter $k_{j}$, and reduces the effect of outliers by introducing relative symmetry. A local average is then computed on the transformation: $|U1[\lambda_{m = 1}]|$ that aggregates the coefficients to build the desired translation-invariant representation:
  
\begin{equation}
\begin{aligned}
S[\lambda_{m = 1}] = |U1[\lambda_{m = 1}]| \star \phi_{2^J}, \\ U1[j] = \log(U[j] + k_{j}), \ U[j] = |x\star \psi_{j}|, 
\end{aligned}
\end{equation}

The above transformation leads to loss of high frequency components due to smoothing. These are thus retrieved by cascades wavelet filtering performed in the second layer. However, these retrieved components are not translation invariant, which is restored by first applying the $L_{2}$ non-linearity of eq($2$) to obtain the regular envelope:

\begin{equation}
U2[\lambda_{m = 1},\lambda_{m = 2}] = |U1[\lambda_{m = 1}] \star \psi_{\lambda_{m = 2}}|
\end{equation}

A local-smoothing operator is then applied to the regular envelope 
($U2[\lambda_{m = 1},\lambda_{m = 2}]$) to extract the desired second layer ($m = 2$) translation invariant coefficients: 
\begin{equation}
S[\lambda_{m = 1},\lambda_{m = 2}] = U2[\lambda_{m = 1},\lambda_{m = 2}] \star \phi_{2^J}
\end{equation}

Finally, the scattering coefficients obtained at each layer are:
\begin{equation}
S = \begin{pmatrix}
x \star \phi_{2^J},
U1[\lambda_{m = 1}] \star \phi_{2^J},
U2[\lambda_{m = 1},\lambda_{m = 2}] \star \phi_{2^J}
\end{pmatrix}
\end{equation} 
\subsubsection{Vesselness}
Traits of ridge-like and tubular structures depicting structural properties can enhance segmentation tasks. The brain matter contains these structures that can be exploited for superior segmentation. In the recent past, these structures have been extracted using eigen-decomposition of the Hessian calculated at each pixel \cite{aylward2002initialization,lindeberg1998edge}. These filters $\xi_j$, for $10$ different scales ($j$), were used to identify narrow vessels in low contrast. The Vesselness filter uses the Hessian matrix and decomposes its eigenvalues into pre-defined (3) orthonormal directions. Scale invariance is obtained using a filter patch of size $10 \times 10$ at different scales \textit{j}, followed by a max operator across the scales to obtain the final estimate of vesselness. This process is presented in Figure~\ref{fig:vesselness}.

\subsubsection{Texture}
Brain Matter typically demonstrates different texture properties for grey and white matter. Grey matter generally present a rougher texture as compared to white matter. In this paper, we use MR8 filter \cite{pitiot2004expert} to achieve rotations and scale invariance while extracting texture information. The filter consists of a Gaussian and a Laplacian of Gaussian filter, an edge filter $\Omega_{j,r}^1$ at 3 scales ($j$) and a bar filter $\Omega_{j,r}^2$ at the same 3 scales ($j$). The later two filters are oriented at 6 directions ($r$). We take the maximum response across orientations to achieve rotational invariance. This reduces the number of filter responses from 38 (6 orientations at 3 scales for 2 oriented filters, plus 2 isotropic) to 8 (3 scales for 2 filters, plus 2 isotropic). Following this, max operation is performed on these features across scales to get scale invariance (similar to invariance in the previous section).

\begin{figure*}[t!]
	\centering
	\includegraphics[width=1\textwidth]{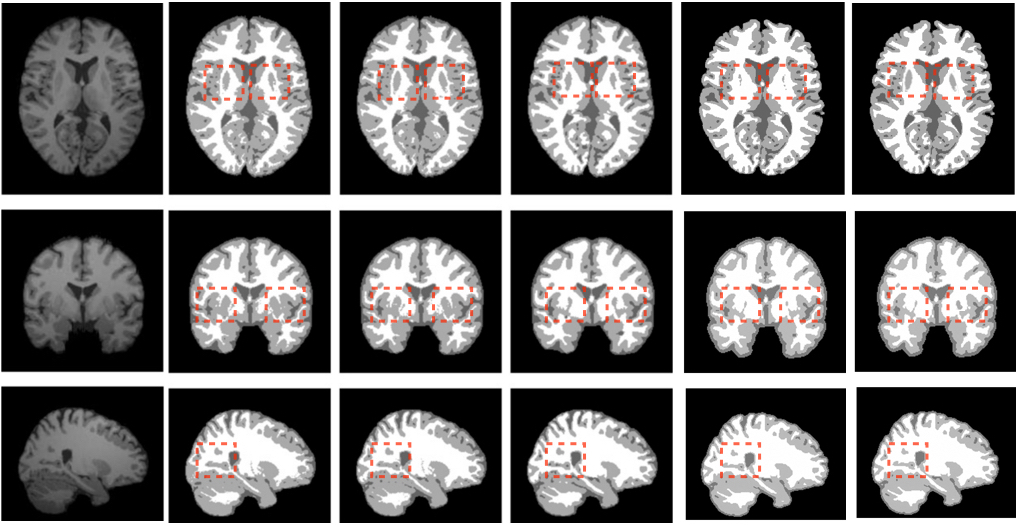}
	\caption{Qualitative Comparison with previous state of the art methods.
	 Column 1: Original images. Column 2:  Ground Truth. Column 3: Leemput et al.'s method~\cite{van1999automated}. Column 4: Wang et al.'s method~\cite{wang2009active}. Column 5: Madiraju et. al.'s method~\cite{madiraju2016level} and Column 6: Proposed TS-SHDL. The regions with the difference are highlighted with red boxes.}
	\label{fig:compare}
\end{figure*}

\subsection{Unsupervised Learning: PCANet}
\label{PCA}
The features extracted using ScatterNet, vesselness, and texture, as described in the previous section, are concatenated and fed to the stacked four-layer PCANet~\cite{pcanet}. The PCANet at each layer learns a translation, rotation, and scale invariant mid-level representation using unsupervised learning from the features of the previous layer. The invariant mid-level features are learned by cropping patches of size $5 \times 5$ from all the feature as $ x_{1}, x_{2},..., x_{mn}$. After this, patch mean is subtracted from each patch to obtain  $\bar{ X} = [\bar{  x}_{1},\bar{ x}_{2},...,\bar{x}_{mn}]$, where $\bar{ x}_{j}$ is a mean-removed patch.

The objective of the layer PCANet is to minimize the reconstruction error within a family of orthonormal filters. This can be seen in the following equation, 
\begin{equation}
\min_{ V\in R^{z_1z_2 \times K}} \| X -  V V^T\ X\|_F^2,~{\rm s.t.}~  V^T V =   I_{K},
\end{equation} 
where,
\begin{equation}\label{eq: datamatrix_1}
 X = [\bar{ X}_1,\bar{ X}_2,...,\bar{ X}_N]\in {R}^{z_1z_2 \times N}.
\end{equation}
$N$ is the number of training images, $ I_{K}$ is identity matrix of size $K\times K$. Here $K$ is the number of filters comprising of the leading $K$ principal eigenvectors capture the main variance amongst the patches for each layer. 

\subsection{Supervised Fisher Encoding}
The Fisher Kernel~\cite{FV},  is used to transform the PCA features obtained from the unsupervised learning module into a compact and dense Fisher vector encoding using supervised learning. The output from the PCA layer is unrolled to get a vector $X = \{x_1, x_2, ...x_T\}$. Assuming a probability density function $u_{\lambda}$ which models the generation process of $X$. The kernel is the described as:
\begin{equation}\label{eq:FisherKernel}
K(X,Y) = G_{\lambda}^{X'}F_{\lambda}^{-1}G_{\lambda}^{Y}
\end{equation}
here $F_{\lambda}$ is the Fisher information matrix, $K$ is the Fisher Kernel on the gradients of log-likelihood. The dimensionality of the Fisher Vector depends on the number of parameters $\lambda$. Further, $F_{\lambda}$ being symmetric and positive definite, can be decomposed as $F_{\lambda} = L_{\lambda}^{'}L_{\lambda}$, which inturn enables the Fisher Kernel to be re-written as the dot product of $H_{\lambda} = L_{\lambda}G_{\lambda}^{X}$. Here  $H_{\lambda}$ is the Fisher Vector corresponding to input $X$.

\subsection{Supervised CRF Segmentation}
\par
The Fisher vector (FV) features are used by the Conditional Random Field (CRF) to achieve the desired brain matter segmentation. CRF is typically a probabilistic framework that establishes a relationship between the pixel labels and extracted features~\cite{domke2013learning}. It is an undirected graphical model that uses 4 pairwise connected grids having finite vertices or nodes and corresponding edges connecting the vertices. Here, each node represents a random variable $X$ and the edges define the neighbourhood relation between these unobserved random variables. To perform the image segmentation, a clique loss function is used with Tree-Reweighted~\cite{domke2013learning} inference which uses LBFGS optimization algorithm. During training, after each iteration, the loss value is checked and if bad search direction is encountered, L-BFGS is reinitialized~\cite{domke2013learning}.

\section{Experimental Results}
This section provides details of the experiments performed on a 2D synthetic datasets (the McGill Brain Web \cite{kwan1999mri}) and a 3D real dataset (MRBrainS website \cite{mendrik2015mrbrains}). The qualitative and quantitative comparison is presented with the state of the art brain segmentation methods on both datasets. The images used in the experimentation have a brain images of intensity
non-uniformity(INU) = 100\% and noise level of 3\%.

\subsection{Synthetic Image Dataset}
A total of 300 brain MRI images are chosen from the McGill Brain Web~\footnote{http://mcgill.ca/bic} \cite{kwan1999mri}. The dataset, split into $140\mid 60\mid 100$ images for $train\mid validation\mid test$ sets, contains realistic MRI images of the human brain produced by an MRI simulator based on two anatomical models: normal and multiple sclerosis (MS). We provide both qualitative (Fig. 1) and quantitative (Table. 1) comparison to evaluate the performance of our proposed method with the state of the art. 

\begin{table*}[!t]
	\centering
	\caption{The table shows comparison of average Jaccard ($\mathcal{J}$) values of grey matter and white matter of our algorithm with current state of the art algorithms on the 2D synthetic MRI brain dataset \cite{kwan1999mri}. Grey: Best Results and Bold: TS-SHDL Results.}
	
	\begin{tabular}{|c||c|c|c|c|c|c|} 
		\hline
		\multirow{2}{*}{Method} &  FSL & SPM & MAP with & Decision Forest  & HDF& TS-SHDL\\
		&~\cite{zhang2001segmentation}&~\cite{ashburner1997multimodal}&histograms~\cite{yi2009discriminative}&Classifier~\cite{zhang2001segmentation}&~\cite{madiraju2016level}& \\
		\hline\hline
		$\mathcal{J}$ of Grey Matter&$0.756^2$&$0.790^2$&$0.814$&$0.838$& $0.89$&~\cellcolor{gray!50}\textbf{0.907}\\
		$\mathcal{J}$ of White Matter&NA&NA&$0.710$&$0.731$& $0.91$&~\cellcolor{gray!50}\textbf{0.921}\\
		\hline
	\end{tabular}
	
	\label{table:1}
\end{table*}

\subsection{Real Image Dataset}
Multisequence 3T MRI scans of twenty people available on the MRBrainS website~\footnote{http://mrbrains13.isi.uu.nl/} \cite{mendrik2015mrbrains} is used as the real image dataset. In the dataset, the subjects were selected to have varying degrees of atrophy and white matter lesions. All the scans have manual segmentations into Grey (GM) and White Matter (WM). 5 subject's scans are used as the train set and the remaining used as test set. We provide extensive quantitative results on this dataset as presented in Table. 2.

\subsection{TS-SHDL Parameters}
This section presents the parameters for the (i) hand-crafted, (ii) unsupervised ,and (iii) supervised block used for the experiments. 

\begin{itemize}

\item \textit{Hand-crafted Module}: The ScatterNet used 2 DTCWT scales, 10 scales were used for the vesselness filter while 3 scales and 6 orientations were used for the texture filter.

\item  \textit{Unsupervised Learning Module}: The number of filter learned in the first, second, third and fourth, PCA-Net layer are 40, 30, 20 and 10, respectively.

\item  \textit{Supervised Learning Module}: The number of GMM cluster for the Fisher encoding are chosen as 5 and the dimension of the final feature vector to be encoded is chosen to be 500. 

\end{itemize}

\subsection{Qualitative Analysis}
\par Figure \ref{fig:compare} provides the qualitative comparison of our proposed method with other state of the art methods. We perform these analyses on the synthetic image dataset. Being 2-dimensional images, it aids visualization and qualitative inspection. Upon close observation, it can be inferred that our method provides superior segmentation and better correlation with ground truth. In the axial image section (first row), it can be seen that our method correctly segments the central grey lobes which was missing in Madiraju et. al.'s method~\cite{madiraju2016level}. Also in the coronal (second row) image, the grey matter segmentation is best represented in our method as compared to the other methods. This shows a qualitative improvement over both grey and white matter segmentation.

\subsection{Quantitative Analysis}\label{sec:quant}
To provide emphasis over the effectiveness of our method, we also provide quantitative evidence using the Jaccard similarity~\cite{vovk2007review} metric on the 2D synthetic brain MRI dataset and more robust Dice coefficient~\cite{cciccek20163d} (DC), $95^{th}$ percentile of Hausdorff distance~\cite{cciccek20163d} (HD), and Absolute Volume Difference~\cite{cciccek20163d} (AVD) metrics on the 3D real brain MRI \cite{mendrik2015mrbrains} dataset for both grey and white matter. 

\subsubsection{2D MRI Synthetic Dataset}
Jaccard similarity is used to evaluate the performance of the proposed TS-SHDL network on the 2D MRI Synthetic Dataset \cite{kwan1999mri}. Jaccard similarity measures the similarity of the segmented regions with the ground truth:
\begin{equation}
\mathcal{J}(G,S)=\frac{\vert G \cap S \vert }{\vert G \cup S \vert}
\end{equation}
value of $\mathcal{J}$ lies between $0$ and $1$. Higher the value of $\mathcal{J}$ accurate the segmentation. 

Table. \ref{table:1} presents an extensive comparison with other brain matter segmentation algorithms applied on the synthetic dataset. Here our method outperforms the state of the art methods in both grey and white matter segmentation. The model also surpasses the recent state of the art method of Madiraju et. al.~\cite{madiraju2016level} thus showing its superior performance.

The proposed TS-SHDL network is also compared with the supervised deep Fully Convolutional Network (FCN)~\cite{cnn} with 8-pixel stride for Brain matter segmentation. The training data fine-tunes the FCN trained on ImageNet and then test on the brain MR image test to achieved a Jaccard index of 0.87 and 0.88 for the grey and white matter respectively. The FCN underperformed because the training data is not sufficient to fine-tune the FCN learned on the ImageNet.

\begin{table*}[!t]
	\centering
	\caption{The table shows comparison of different methods (DC: \%, HD: mm, AVD: \%) of grey matter and white matter of our algorithm with current state of the art algorithms on the 3D real MRI brain dataset \cite{mendrik2015mrbrains}. Grey: Best Results and Bold: TS-SHDL Results.}
	
	\begin{tabular}{|cc||c|c|c|c|c|c|} 
		\hline
		\multicolumn{1}{|c}{\multirow{2}{*}{Method}}& &  3D U-net & PyraMID & VoxResNet & Mahbod et. al.  & Pereira et. al.& TS-SHDL\\
		&&~\cite{cciccek20163d}&~\cite{stollenga2015parallel}&\cite{chen2016voxresnet}&\cite{mahbod2016structural}&~\cite{pereira2016automatic}& \\
		
		\hline\hline
		
		\multirow{3}{*}{Grey Matter}&DC&85.44&84.82&\cellcolor{gray!50}86.15&84.77&84.50&~ \textbf{86.01}\\
		&HD&1.58&1.70&1.45&1.71&1.70& \cellcolor{gray!50}\textbf{1.40} \\
		&AVD&6.60&6.77&6.60&6.02&7.10& \cellcolor{gray!50}\textbf{6.46} \\
		\hline
		\multirow{3}{*}{White Matter}&DC&88.86&88.33&89.46&88.45&88.04&~\cellcolor{gray!50}\textbf{89.49}\\
		&HD&1.95&2.08&1.90&2.34&2.12& \cellcolor{gray!50}\textbf{1.88}\\
		&AVD&6.47&7.05&6.05&7.67&7.74&\cellcolor{gray!50}\textbf{5.99} \\
		\hline
	\end{tabular}
	\label{table:2}
\end{table*}

\subsubsection{3D MRI Real Dataset}
The performance of the proposed TS-SHDL is evaluated on the 3D MRI Real Dataset \cite{mendrik2015mrbrains} using three robust evaluations metrics: (i) Dice coefficient~\cite{cciccek20163d}, (ii) Hausdorff distance~\cite{cciccek20163d}, and (iii) Absolute Volume Difference~\cite{cciccek20163d}.

Dice coefficient~\cite{cciccek20163d} measures the spatial overlap between the ground truth and the predicted segmentation. It is defined as, 
\begin{equation}
D(G,S)=\frac{2\vert G \cap S \vert }{\vert G \vert + \vert S \vert}. 100\%
\end{equation}
Here, $G$ is the ground truth while $S$ is the segmentation. A higher value of the Dice coefficient signifies higher segmentation accuracy.

Hausdorff distance~\cite{cciccek20163d} measures the distance between segmentation results and ground truth. In-order to be insensitive to outliers, we use the $K^{th}$ ranked distance, 
\begin{gather}
h_{95}(S, G) =    ^{95}K_{s \in S}^{th} \quad \underset{g \in G}{min} \vert \vert g - s \vert \vert \\
HD(G,S)= max\{ h_{95}(S, G), h_{95}(G, S) \}
\end{gather}  
Unlike Dice coefficient,  a smaller $HD$ denotes higher accuracy. 

Finally, we calculate Absolute Volume Difference~\cite{cciccek20163d} which is defined as, 
\begin{equation}
AVD(G,S)=\frac{\vert V_s - V_g \vert }{V_g}. 100\%
\end{equation} 
where, $V_s, V_g$ are the  volume of segmentation results and ground truth. Similar to $HD$, lesser value of $AVD$ signifies better accuracy.
All the three metrics are applied for both grey and white matter in the real image dataset.

We have compared our method with both state of the art deeplearning based models that include 3D U-net \cite{cciccek20163d}, PyraMID-LSTM \cite{stollenga2015parallel}, VoxResNet \cite{chen2016voxresnet} and hand-crafted feature based models such as \cite{mahbod2016structural, pereira2016automatic} etc. 3D U-net extracts volumetric feature representation from the 3D images, PyraMID paralellises multi-dimensional RNNs in a pyramid fashion and VoxResNet provides a voxelwise residual network for performing the segmentation. On the other hand, methods based on hand-crafted features use different techniques that involve histogram based features \cite{mahbod2016structural} and gradients \cite{pereira2016automatic} etc. As seen in Table \ref{table:2}, our model is able to outperform all these methods in the real image dataset.

\section{Conclusion}
The paper presents the ScatterNet Hybrid Deep Learning Network (SHDL) with Texture and Structure Features that capture the intensity and structural properties of the brain matter. Four layers of the PCANet are used to learn hierarchical invariant mid-level features from the concatenated hand-crafted features. Fisher Encoding is used on the mid-level feature to obtain a compact and dense representation that is finally used by the CRF is used to achieve the desired brain matter segmentation. The proposed network outperformed the state-of-the-art on both qualitative and quantitative measures. In addition, the proposed network can be the potential model choice for applications with small datasets as deep supervised learning network may not be trainable because of insufficient the training dataset. 
{\small
\bibliographystyle{ieee}
\bibliography{egbib}
}

\end{document}